%% file: main.tex
\pdfoutput=1
\documentclass[11pt]{article}

\usepackage{acl}
\usepackage{times}
\usepackage{latexsym}
\usepackage[T1]{fontenc}
\usepackage[utf8]{inputenc}

\usepackage{microtype}
\usepackage{enumitem}

\usepackage{hyperref}
\usepackage{booktabs}
\usepackage{graphicx}
\graphicspath{{./imgs/}}
\usepackage{footmisc}
\usepackage{microtype}
\usepackage{arydshln}
\usepackage{caption}
\usepackage{subcaption}
\usepackage{array, makecell} %
\usepackage{footmisc}
\usepackage{alltt}
\usepackage{floatrow}
\usepackage{pifont}% http://ctan.org/pkg/pifont

\usepackage{tabularx}
\usepackage{adjustbox}
\usepackage{multirow}

\usepackage[table, dvipsnames]{xcolor} % For coloring table cells
\usepackage{multirow, colortbl, caption}
\definecolor{lightblue}{RGB}{232, 244, 248}
\definecolor{lightpink}{RGB}{254, 238, 237}

\definecolor{bluelink}{RGB}{0,113,188}
\definecolor{greenlink}{RGB}{0,188,113}
\definecolor{darkblue}{rgb}{0,0,0.55} %

\usepackage{tikz}

\usepackage{collcell}

\usepackage{etoolbox}

\newtoggle{inTableHeader}% Track if still in header of table
\toggletrue{inTableHeader}% Set initial value
\newcommand*{\StartTableHeader}{\global\toggletrue{inTableHeader}}%
\let\OldTabular\tabular%
\let\OldEndTabular\endtabular%
\renewenvironment{tabular}{\StartTableHeader\OldTabular}{\OldEndTabular\StartTableHeader}%

\newcommand*{\MinNumber}{-1.0}%
\newcommand*{\MidNumber}{0.0} %
\newcommand*{\MaxNumber}{1.0}%

\newcommand{\ApplyGradient}[1]{%
  \iftoggle{inTableHeader}{#1}{
    \ifdim #1 pt > \MidNumber pt
        \pgfmathsetmacro{\PercentColor}{max(min(100.0*(#1 - \MidNumber)/(\MaxNumber-\MidNumber),100.0),0.00)} %
        \hspace{-0.33em}\colorbox{yellow!\PercentColor!blue}{#1}
    \else
        \pgfmathsetmacro{\PercentColor}{max(min(100.0*(\MidNumber - #1)/(\MidNumber-\MinNumber),100.0),0.00)} %
        \hspace{-0.33em}\colorbox{blue!\PercentColor!blue}{#1}
    \fi
  }}
\newcolumntype{R}{>{\collectcell\ApplyGradient}c<{\endcollectcell}}

\input{math-com}

\usepackage[nameinlink]{cleveref}
\crefformat{section}{\S#2#1#3} % see manual of cleveref, section 8.2.1
\crefname{algorithm}{Alg.}{Algs.}
\crefname{table}{Table}{Tables}
\crefformat{subsection}{\S#2#1#3}
\Crefname{equation}{Eq.}{Eqs.}
\Crefname{figure}{Figure}{Figures}

\usepackage[colorinlistoftodos,prependcaption,textsize=tiny]{todonotes}

\definecolor{darkgreen}{rgb}{0,0.5,0}  
\usepackage{soul}

\usepackage{float}

\usepackage{multirow}% http://ctan.org/pkg/multirow
\usepackage{hhline}% http://ctan.org/pkg/hhline

\definecolor{chartqapro1}{RGB}{30,160,220} % Adjust based on the gradient color (blue)
\definecolor{chartqapro2}{RGB}{50,200,100} % Adjust based on the gradient color (green)

\title{
\textbf{Aligning Text, Code, and Vision: A Multi-Objective Reinforcement Learning Framework for Text-to-Visualization}}

\author{
\textbf{Mizanur Rahman}\textsuperscript{\textdaggerdbl}\thanks{\hspace{0.115cm} Contact Emails: \{mizanurr,enamulh\}@yorku.ca},
\textbf{Mohammed Saidul Islam}\textsuperscript{\textdaggerdbl},
\textbf{Md Tahmid Rahman Laskar}\textsuperscript{\textdaggerdbl} \\
\textbf{Shafiq Joty}\textsuperscript{\textparagraph,\textsection},
\textbf{Enamul Hoque}\textsuperscript{\textdaggerdbl}\footnotemark[1] \\
{\textsuperscript{\textdaggerdbl}York University,
\textsuperscript{\textparagraph}Salesforce AI Research, \quad
\textsuperscript{\textsection}Nanyang Technological University}
}

\begin{document}
\maketitle

\begin{abstract}

Text-to-Visualization (Text2Vis) systems translate natural language queries over tabular data into concise answers and executable visualizations. While closed-source LLMs generate functional code, the resulting charts often lack semantic alignment and clarity—qualities that can only be assessed post-execution. Open-source models struggle even more, frequently producing non-executable or visually poor outputs. Although supervised fine-tuning can improve code executability, it fails to enhance overall visualization quality, as traditional SFT loss cannot capture post-execution feedback. To address this gap, we propose RL-Text2Vis, the first reinforcement learning framework for Text2Vis generation. Built on Group Relative Policy Optimization (GRPO), our method uses a novel multi-objective reward that jointly optimizes textual accuracy, code validity, and visualization quality using post-execution feedback. By training Qwen2.5 models (7B and 14B), RL-Text2Vis achieves a 22\% relative improvement in chart quality over GPT-4o on the Text2Vis benchmark and boosts code execution success from 78\% to 97\% relative to its zero‑shot baseline. Our models significantly outperform strong zero-shot and supervised baselines and also demonstrate robust generalization to out-of-domain datasets like VIS-Eval and NVBench. These results establish GRPO as an effective strategy for structured, multimodal reasoning in visualization generation. We release our code at \url{https://github.com/vis-nlp/RL-Text2Vis}. 

%While closed-source large language models perform reasonably well for code generation, their visualizations often lack semantic alignment and clarity, qualities that can only be assessed after code execution. Open-source models, particularly small to mid-sized ones (7B–14B), which are essential for privacy-sensitive and industrial applications, perform even worse, frequently producing non-executable or visually poor outputs. Supervised fine-tuning improves code executability but fails to enhance visualization quality, as traditional loss functions cannot capture post-execution semantics. 

%To address this gap, we propose RL-Text2Vis, the first reinforcement learning framework for text-to-visualization generation, built on Group Relative Policy Optimization (GRPO) with a novel multi-objective reward that jointly optimizes textual accuracy, code validity, and visualization quality using post-execution feedback. Trained on Qwen2.5 models (7B and 14B), RL-Text2Vis achieves a 22\% relative improvement in chart clarity and correctness over GPT-4o on the Text2Vis, increases code execution success from 78\% to 97\%, and outperforms strong zero-shot and supervised baselines. Our models also generalize strongly to out-of-domain datasets such as VIS-Eval and NVBench, demonstrating robustness across diverse schemas. These results suggest GRPO as an effective strategy for structured multimodal reasoning in visualization generation. We release RL-Text2Vis code at <redacted>.
\end{abstract}

\section{Introduction}

% Data visualization, the graphical representation of information, is central to understanding complex data, identifying patterns, and supporting data-driven decision-making \cite{aparicio2015data, hoque2022chartquestionansweringstate}. However, creating accurate and interpretable visualizations based on user requirements is a multi-step process that involves understanding a natural language query, retrieving and transforming relevant data, and generating the appropriate visualization \cite{shen2022towards}. This process requires programming expertise, familiarity with visualization libraries, and knowledge of design principles, creating a significant barrier for non-technical users. Alongside visualizations, concise textual answers are {often} crucial for quickly validating results and providing context, as users often seek both a summary insight and a visual representation \cite{text2vis2025}. For example, in Figure \ref{fig:performance_gap}, the textual answer “2020” immediately identifies the correct year with the greatest increase in renewables share, helping users interpret the chart without manually extracting values. Text-to-Visualization (Text2Vis)  addresses this challenge by automatically converting natural language queries over tabular data into concise answers and accurate visualizations, thereby eliminating the need for manual coding.   

\begin{figure}[t!]
    \includegraphics[width=\textwidth]{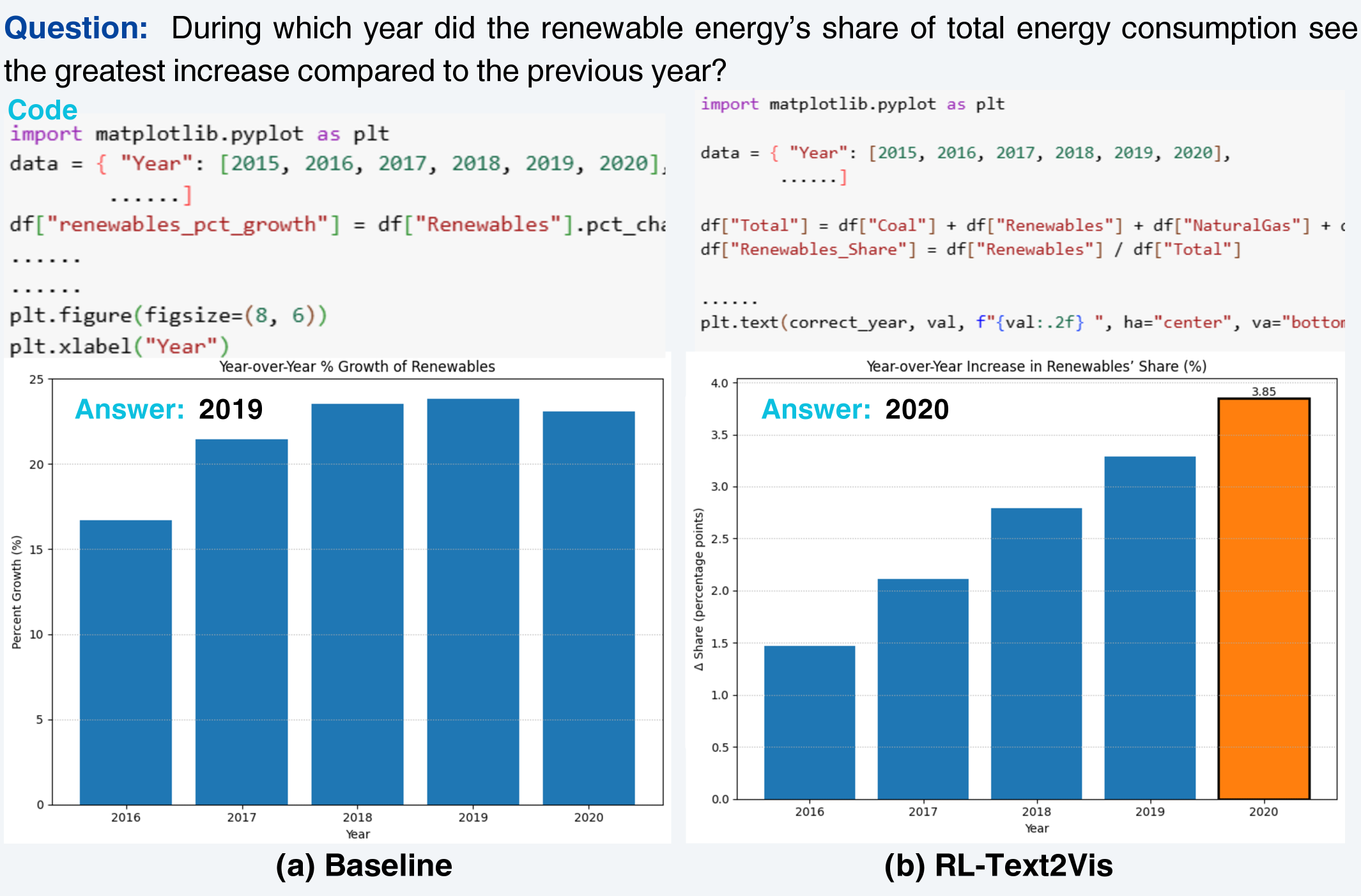}
    \caption{An example from Text2Vis \cite{text2vis2025}. Given the data table and question as input, (a) the baseline model, Qwen2.5-14B-Instruct generates runnable visualization code, but the visualization is not aligned with the query as it shows the growth of renewables quantity instead of the share of total, leading to an incorrect answer. (b) Our RL-Text2Vis-14B produces a correct, query-aligned, and interpretable visualization.
    %Our model RL-Text2Vis-14B produces a correct and interpretable visualization aligned with the input query.
    % \enamul{caption is a bit unclear: what does it mean by answering "wrong question". why don't we show the code snippets of two versions like Fig 3. }
    % \enamul{we should not use the same example from the previous paper. It feels like increamental work. Can we show o/p from GRPO that is correct but the baseline is not correct?  Use 2-col fig if necessary.} \joty{good point. Also avoid self-citing.}
    }
    \label{fig:performance_gap}
    \vspace{-2mm}
    % \vskip -2ex
\end{figure}

Data visualization is central to understanding complex data, identifying patterns, and supporting data-driven decision-making \cite{rahman2025llm, aparicio2015data, hoque2022chartquestionansweringstate}.
However, creating accurate and interpretable visualizations from natural language queries requires programming expertise, familiarity with visualization libraries, and design knowledge—posing a barrier for non-technical users \cite{shen2022towards}. 
%However, creating accurate and interpretable visualizations based on user requirements is a multi-step process that involves understanding a natural language query, retrieving and transforming relevant data, and generating the appropriate visualization \cite{shen2022towards}. This process requires programming expertise, familiarity with visualization libraries, and knowledge of design principles, creating a significant barrier for non-technical users. 
Alongside visualizations, concise textual answers are {often} crucial for quickly validating results and providing context, as users often seek both a summary insight and a visual representation \cite{text2vis2025}. For example, in Figure \ref{fig:performance_gap}, the textual answer “2020” identifies the correct year with the greatest increase in renewables share, helping users interpret the chart without manually extracting values. Text-to-Visualization (Text2Vis)  addresses this challenge by automatically converting natural language queries over tabular data into concise answers and accurate visualizations, thereby eliminating the need for manual coding.   

Despite advances in large language models (LLMs), their ability to generate high-quality visualizations remains limited. Benchmarks such as VisEval \cite{chen2024viseval} and Text2Vis \cite{text2vis2025} reveal persistent failures in semantic correctness, code executability, and chart readability. Closed-source models like GPT-4o, though strong in code generation, often produce visually misaligned charts and raise compliance concerns. Open-source models perform even worse; small to mid-size versions (7B-14B) favored for on-premise deployments frequently produce non-executable or misaligned outputs, while larger models achieve only moderate performance and remain too expensive for many real-world applications \cite{mallick2024chatvis, chen2024viseval}.

Unlike code generation, where aspects of functional correctness can often be verified by execution success (i.e., the program runs without exceptions), Text2Vis introduces additional quality dimensions that are only observable after rendering, notably chart readability, semantic alignment with the query, and visual clarity. Current approaches typically use supervised fine-tuning (SFT) that minimizes token-level loss to imitate reference outputs. While this improves executability by aligning syntax and structure, it cannot optimize these post-execution qualities because traditional loss functions provide no signal about visualization clarity or semantic alignment. This limitation highlights the need for alternative training strategies that incorporate post-execution feedback \cite{wei2025swe}.

% Unlike code generation, where correctness can be verified by execution success, visualization quality depends on aspects such as chart readability, semantic alignment, and visual clarity that can only be assessed after execution. This challenge highlights why supervised fine-tuning, while improving code executability, fails to enhance visualization quality because traditional loss functions cannot capture these post-execution semantics \cite{wei2025swe}.

% Furthermore, supervised fine-tuning can enhance code executability but struggles with visualization quality and alignment because these aspects are inherently post-execution and subjective \cite{wei2025swe}.  

Meanwhile, Reinforcement learning (RL) has emerged as a powerful technique for improving reasoning in LLMs, with models like DeepSeek-R1 \cite{guo2025deepseek} and methods like SWE-RL \cite{wei2025swe} demonstrating significant gains on coding and software engineering tasks. However, these methods primarily rely on single-modal feedback such as numeric correctness or code executability. Data visualization presents a more complex challenge, as its quality depends on a combination of textual accuracy, query alignment, code validity, and the visual clarity of the rendered chart. Effectively optimizing for these multimodal requirements necessitates a post-execution, multi-objective reward that can jointly align the model's reasoning, code, and final visual output.

%reinforcement learning (RL) has emerged as a powerful approach for improving reasoning and structured outputs in LLMs \cite{wang2024reinforcement, xie2025logic}. For instance, DeepSeek-R1 \cite{guo2025deepseek} demonstrated that RL with rule-based rewards can substantially enhance performance on mathematical and coding tasks, while SWE-RL \cite{wei2025swe} applied RL to real-world software engineering challenges. They do not address the multimodal requirements of visualization, which include textual accuracy, query alignment, code validity, and the clarity of the rendered chart. Since visualization quality can only be assessed after execution, effective optimization requires a post‑execution, multi-objective reward that jointly aligns reasoning, code intent, and visual output\cite{text2vis2025}.

In this work, we propose RL-Text2Vis, the first RL framework tailored for the Text2Vis tasks. It uses Group Relative Policy Optimization (GRPO) \cite{wang2023math} to incorporate post-execution feedback, enabling optimization beyond code correctness to include visualization clarity and semantic alignment. Unlike existing RL-based methods that rely on single-modal signals, RL-Text2Vis introduces a novel multi-objective reward that jointly considers three aspects critical for visualization: (i) syntactic and functional validity of the code, (ii) alignment, correctness, and quality of the generated chart, and (iii) textual correctness of the predicted answer. This design explicitly captures the multimodal nature of the problem, addresses the shortcomings of SFT and single-modal RL methods, and improves generalization across diverse data contexts and query types.

 We implement RL-Text2Vis on top of Qwen2.5 Instruct models of size  7B and 14B \cite{yang2024qwen2}, and train them using GRPO with our proposed multi-objective reward. We conduct extensive experiments on the Text2Vis benchmark \cite{text2vis2025} and further evaluate generalization on VIS-Eval \cite{chen2024viseval} and NVBench \cite{luo2021nvbench}. Our RL-Text2Vis outperforms SFT baselines, zero-shot open-source models, and even proprietary systems like GPT‑4o, achieving over 22\% relative improvement compared to GPT‑4o in both chart clarity and correctness. These results demonstrate that RL with post-execution visual feedback is an effective strategy for structured, multimodal reasoning in visualization generation. Moreover, RL-Text2Vis offers a practical, deployable approach for real-world scenarios where privacy, cost, and compliance constraints limit the use of closed-source models. 

In summary, our contributions are:
\textbf{\Ni} We propose RL-Text2Vis, the first RL framework for Text-to-Visualization, optimizing query-aligned and interpretable visualizations. \textbf{\Nii} We introduce a novel multi-objective reward leveraging post-execution feedback to jointly optimize visualization clarity and alignment, code validity, and textual correctness. \textbf{\Niii} Our approach significantly outperforms all baselines, achieving 22\% higher chart clarity and correctness than GPT‑4o and improving code executability to 97\% on the Text2Vis benchmark. \textbf{\Niv} RL-Text2Vis demonstrates strong out-of-domain generalization on VIS-Eval and NVBench, proving robustness across diverse queries and domains. 
%\textbf{\Nv} We provide a practical, deployable solution for real-world scenarios where privacy, cost, and compliance constraints limit the use of closed-source models.

% \vspace{-2mm}

\section{Related Work}

\noindent \textbf{LLMs for Automated Visualization} Early systems for text-to-visualization generation, such as NL4DV \cite{narechania2020nl4dv} and Advisor \cite{liu2021advisor}, relied on rule-based grammars or template-driven specifications. While these approaches ensured syntactic correctness, they were rigid, lacked scalability, and could not handle the diversity of real-world analytical queries. Recent advances in LLMs have enabled more flexible approaches for text-to-visualization generation. Methods such as Chat2VIS \cite{maddigan2023chat2vis}, LIDA \cite{dibia2023lida}, ChartGPT \cite{tian2024chartgpt}, and ChatVis \cite{mallick2024chatvis} translate natural language queries into executable visualization code through prompting, while Prompt4Vis \cite{li2024prompt4vis} enhances schema-aware prompting for improved accuracy. ChartLlama \cite{han2023chartllama} leverages instruction tuning to improve chart reasoning and generation quality. Frameworks like MDSF \cite{zhang2025mdsf} extend this paradigm to multi-dimensional data storytelling by integrating contextual insights with automated visualization.

Despite these advances, LLM-based methods still face critical challenges. They often produce incorrect or incomplete answers, fail to generate executable code, or create charts that lack clarity or relevance to the analytical intent \cite{chen2024viseval, text2vis2025}. These limitations indicate that supervised fine-tuning and prompt engineering alone are insufficient: SFT optimizes only token-level likelihoods, often improving   executability via syntax/library alignment. However, it cannot target post-execution qualities such as visual clarity. Prompting, by design, does not update model parameters \cite{zhang2024instructiontuninglargelanguage}. Consequently, neither incorporates the multimodal feedback available after execution that is needed for interpretable, trustworthy visualizations. We address this by optimizing a post-execution, multi-objective reward.   
% , as they do not incorporate post-execution evaluation or multimodal optimization, both of which are essential for generating interpretable and trustworthy visualizations.
 
\noindent \textbf{Visualization Evaluation} Evaluating visualization generation has traditionally relied on rule-based metrics, which fail to capture semantic alignment or visual quality, or on manual evaluation, which is labor-intensive and cannot scale \cite{liu2021advisor, srinivasan2021collecting}. Recent work shows that LLMs and vision-language models (VLMs) can serve as reliable evaluators by reasoning jointly over text, code, and images \cite{chen2024viseval,text2vis2025}. Closed-source models such as GPT-4o and Gemini \cite{gu2024survey} provide strong evaluation capabilities, while open-source alternatives like Qwen-VL and LLaVA offer practical solutions for privacy-constrained environments \cite{laskar2025judging}. These models assess both textual correctness and visual quality, making them suitable as reward models. However, existing evaluation methods are primarily used for post-hoc benchmarking rather than model optimization. We integrate multimodal, post‑execution feedback (text, code, image) directly into the training loop by converting LLM/VLM evaluators into a multi‑objective RL reward.

\noindent \textbf{RL for Code Generation and Reasoning} RL has emerged as a powerful approach for improving reasoning and structured outputs in LLMs \cite{tie2025survey}. Techniques such as RLHF \citep{ouyang2022training} and Direct Preference Optimization (DPO) \citep{rafailov2023direct} align models with human preferences with respect to safety, factual correctness and stylistic aspects such as formats rather than structured reasoning or execution correctness. Proximal Policy Optimization (PPO) based approaches \citep{schulman2017proximal} use rule-based rewards (e.g., execution success, exact answer matching) but require a critic model, making them computationally expensive. %and unstable under sparse or multimodal rewards.

Recent work shows RL from verifiable rewards improves LLM reasoning. DeepSeek-R1 \citep{guo2025deepseek} introduced GRPO, a scalable policy-gradient method that removes the learned critic by ranking multiple outputs per prompt and computing relative advantages. SWE-RL \citep{wei2025swe} applied GRPO to software engineering tasks, 
showing strong results in structured, code-intensive domains, yet these methods rely on single-modal feedback such as code execution success or numeric correctness. CodeRL \citep{le2022coderl} optimizes code synthesis via functional correctness and unit-test rewards, while ChartGPT \citep{tian2024chartgpt} addresses chart generation through supervised or heuristic feedback.

Text-to-Visualization requires optimizing multiple post-execution objectives: semantic correctness and query alignment, code validity and executability, and chart readability. To our knowledge, no prior work integrates post-execution multimodal feedback into RL or combines it with a multi-objective reward balancing textual, code, and visualization quality. Our RL-Text2Vis framework uses GRPO with this dual novelty, enabling scalable optimization across all visualization dimensions without training a value network.

\begin{figure*}[t]
    \vspace{-5mm}
    \centering    \includegraphics[width=\textwidth]{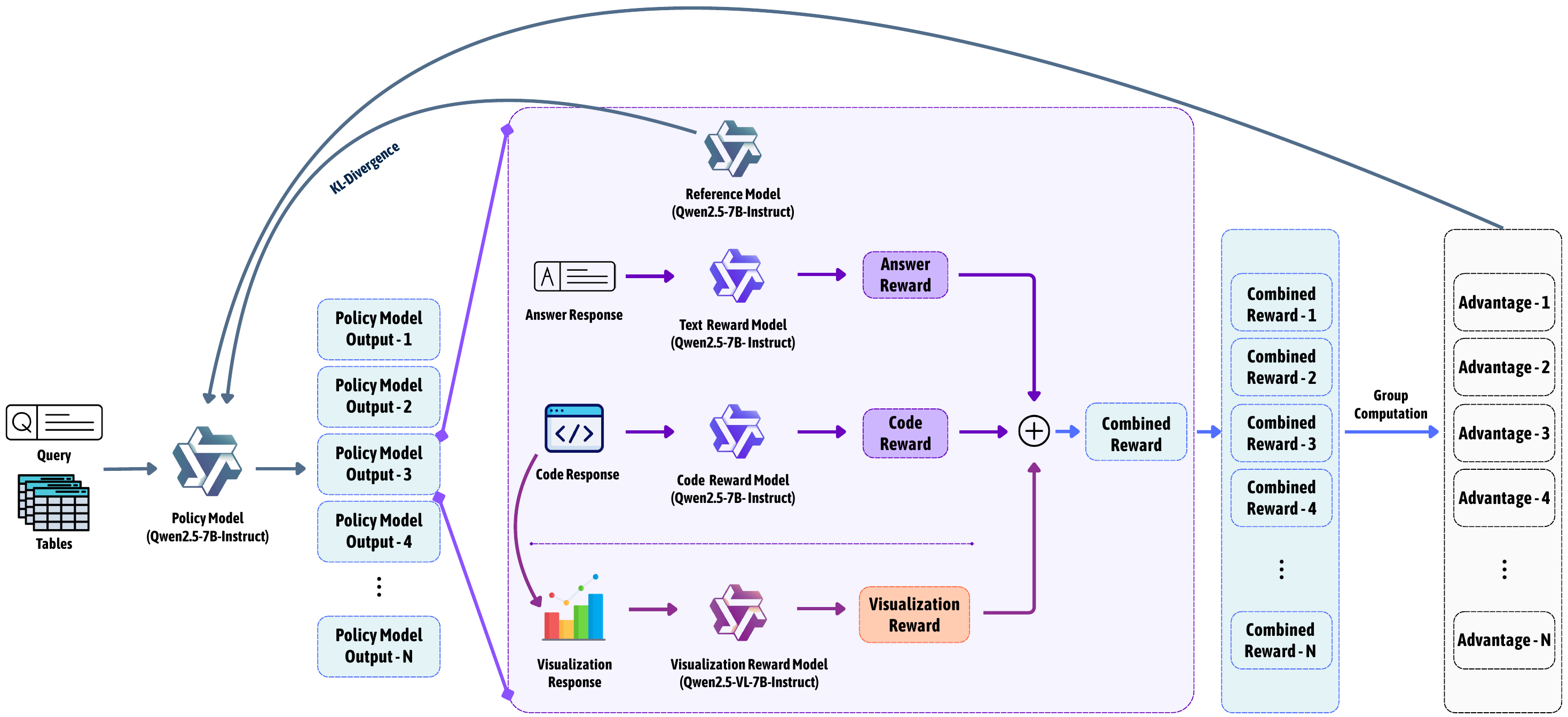}
    \caption{GRPO training architecture for text-to-visualization generation, showing policy outputs, multi-objective rewards (answer, code, visualization), combined reward computation, and advantage calculation for policy updates.}
    % \caption{Overview of our RL training methodology.}
    \label{fig:method}
    \vspace{-3mm}
    \vskip -.5ex
\end{figure*}

% \vspace{-2mm}

\section{\textsc{RL-Text2Vis}}
\input{Dataset}

\section{Conclusion}
We introduced RL-Text2Vis, the first reinforcement learning framework for text-to-visualization generation that integrates multimodal, post-execution feedback into training. Unlike supervised approaches that optimize token-level likelihoods, RL-Text2Vis jointly optimizes textual accuracy, code executability, and visualization quality through GRPO with a multi-objective reward. Experiments show consistent gains across models and benchmarks, demonstrating that reinforcement learning with post-execution visual feedback enhances structured and multimodal reasoning. 
The framework matches or surpasses proprietary models like GPT-4o while being open, 
efficient, and privacy-preserving, highlighting the %broader
potential of multi-objective RL for advancing multimodal generation.

\section*{Ethical Considerations}

RL-Text2Vis is designed to improve transparency in text-to-visualization generation; however, ethical risks remain. Visualizations can amplify biases present in the data or queries, potentially leading to misinterpretation in sensitive domains like healthcare or finance. While our benchmarks use synthetic or public datasets to avoid privacy issues, real-world deployments must ensure data anonymization and human oversight. Additionally, the computational cost of RL raises environmental concerns, which we mitigate through model reuse and efficient training strategies. Finally, safeguards should be implemented to prevent misuse, such as generating deceptive visualizations or applying the system in adversarial contexts.

\section*{Limitations}

Although RL-Text2Vis demonstrates significant improvements in text-to-visualization generation, several limitations remain. First, the 14B model yields the best quality but requires considerably more computation and memory. For resource‑constrained organizations, the 7B variant is a more practical and cost‑effective choice. Second, we did not train larger backbones (e.g., 32B or 72B) due to resource constraints, but our scaling from 3B→7B→14B already shows consistent gains, suggesting that further improvements are likely at larger scales, which remains a promising direction for future work. Third, while our method generalizes to out-of-domain benchmarks, its robustness in highly specialized domains (e.g., medical or financial visualizations) remains untested. Fourth, our study focuses on static visualizations; extending this framework to interactive or multi-view visual analytics is an important direction for future work.
% Finally, our study focuses on static visualizations; extending this framework to interactive or multi-view visual analytics is an important direction for future work. We also did not explore larger GRPO group sizes (e.g., 16 or 32 candidates per prompt); investigating such hyperparameters could yield further gains but is currently constrained by computational cost, which we leave for future work.

\bibliographystyle{acl_natbib}
\bibliography{text2vis}
\newpage
\input{Appendix}

\end{document}

%% file: math-com.tex
\usepackage{amsmath}
\usepackage{amsfonts,bm}
\usepackage{xspace}

\newcommand{\Ni}{({\em i})~}
\newcommand{\Nii}{({\em ii})~}
\newcommand{\Niii}{({\em iii})~}
\newcommand{\Niv}{({\em iv})~}

\definecolor{mypink3}{cmyk}{0, 0.7808, 0.4429, 0.1412}

\makeatletter   
\newcommand{\sveryshortarrow}[1][3pt]{\mathrel{%
    \vcenter{\hbox{\rule[-.5\fontdimen8\scriptfont3]
               {\scriptratio\dimexpr#1\relax}{\fontdimen8\scriptfont3}}}%
   \mkern-4mu\hbox{\let\f@size\sf@size\usefont{U}{lasy}{m}{n}\symbol{41}}}}
\makeatother

\def\eqref#1{equation~\ref{#1}}
\def\1{\bm{1}}

\def\m1{{\bm{1}}}

\DeclareMathAlphabet{\mathsfit}{\encodingdefault}{\sfdefault}{m}{sl}
\SetMathAlphabet{\mathsfit}{bold}{\encodingdefault}{\sfdefault}{bx}{n}

%% file: Dataset.tex
%\vspace{-2mm}

We introduce \textbf{RL-Text2Vis}, the first RL framework for Text2Vis generation that leverages post-execution, multimodal feedback, since visualization quality is only observable after rendering. 
% This structured feedback across multiple modalities improves semantic accuracy, code executability, and visualization quality.
The pipeline, illustrated in Figure~\ref{fig:method}, operates as follows: given a natural language query and a table, the policy produces a structured output containing a concise answer and visualization code. Each output is scored by a two-stage reward: (1) a format reward enforcing structural validity and (2) a composite multimodal reward. Policy updates are performed with GRPO, which provides efficient training without a learned critic model, while directly aligning the model to high-level quality objectives.

%\subsection{Overview} 

\vspace{-2mm}

\subsection{Problem Formulation}

We formalize Text2Vis generation as mapping an input $x = (\text{query}, \text{table}) \in \mathcal{D}$ to an output $y = (\text{answer}, \text{code}) \sim \pi_\theta(\cdot \mid x)$, where $\pi_\theta$ denotes the policy with parameters $\theta$. After executing the generated code to render a chart, a scalar reward $R(x, y)$ is assigned using a two-stage mechanism that combines structural and quality-based signals (see Section~\ref{subsec:reward}). Under this RL formulation, the objective is to maximize the expected reward:
\begin{equation}
\max_\theta \; \mathbb{E}_{x \sim \mathcal{D}, \; y \sim \pi_\theta(\cdot \mid x)} \big[ R(x, y) \big],
\label{eq:rl-objective}
\end{equation}
Unlike SFT, which minimizes the negative log-likelihood of reference outputs (i.e., off-policy), this objective is on-policy and directly targets multi-objective, post-execution criteria such as visualization clarity, semantic alignment, and code executability that cannot be expressed through token-level losses. 

% This enables direct improvement in visualization interpretability and correctness beyond what supervised methods can achieve.

% We formalize text-to-visualization generation as mapping an input $x = {\text{query}, \text{table}}$ sampled from a dataset $\mathcal{D}$ to a response $y = {\text{answer}, \text{code}}$ generated by a model $\pi_\theta$, where $\theta$ denotes the model parameters. We formulate this as a conditional generation problem under a reinforcement learning framework, where the objective is to maximize the expected reward:
% \begin{equation}
% \max_\theta \; \mathbb{E}_{x \sim \mathcal{D}, \; y \sim \pi_\theta(\cdot|x)} \big[ R(x, y) \big],
% \label{eq:rl-objective}
% \end{equation}
% where $R(x, y)$ is a two-stage reward that combines structural and quality-based signals (detailed in Section~\ref{subsec:reward}). Unlike supervised fine-tuning, which minimizes negative log-likelihood over reference outputs, this formulation enables optimization with respect to multi-objective metrics beyond text similarity. 
\vspace{-2mm}

\subsection{GRPO Optimization} 

% \begin{figure*}[t]
% \small
% \begin{equation}
% \mathcal{J}_{\text{GRPO}}(\theta)=
% \mathbb{E}\Bigg[
% \frac{1}{G}\sum_{i=1}^{G}\;
% \mathcal{T}_i
% \Bigg(
% \min\!\Bigg(
% \frac{\pi_\theta\!\bigl(o_{i,t}\,|\,q,o_{i,<t}\bigr)}
%      {\pi_{\text{old}}\!\bigl(o_{i,t}\,|\,q,o_{i,<t}\bigr)}
% \,\hat{A}_{i,t},
% \;
% \text{clip}\!\Bigg(
% \frac{\pi_\theta\!\bigl(o_{i,t}\,|\,q,o_{i,<t}\bigr)}
%      {\pi_{\text{old}}\!\bigl(o_{i,t}\,|\,q,o_{i,<t}\bigr)},\;
% 1-\varepsilon,\;1+\varepsilon
% \Bigg)
% \hat{A}_{i,t}
% \Bigg)
% -\beta\,D_{\mathrm{KL}}\!\bigl[\pi_\theta \,\|\, \pi_{\text{ref}}\bigr]
% \Bigg)
% \Bigg],
% \tag{3}
% \label{fig:grpo_compact}
% \end{equation}
% \end{figure*}

% \begin{figure*}[t]
% \small
% \begin{equation}
% \mathcal{J}_{\text{GRPO}}(\theta)=
% \mathbb{E}\Bigg[
% \frac{1}{G}\sum_{i=1}^{G}\;
% \frac{1}{|o_i|} \sum_{t=1}^{|o_i|}
% \Bigg(
% \min\!\Bigg(
% i_t(\theta)
% \,\hat{A}_{i,t},
% \;
% \text{clip}\!\Big(
% i_t(\theta),\;
% 1-\varepsilon,\;1+\varepsilon
% \Big)
% \hat{A}_{i,t}
% \Bigg)
% -\beta\,D_{\mathrm{KL}}\!\bigl[\pi_\theta \,\|\, \pi_{\text{ref}}\bigr]
% \Bigg)
% \Bigg],
% \tag{3}
% \label{fig:grpo_compact}
% \end{equation}
% \end{figure*}

{
\small
\begin{equation}
\begin{aligned}
\mathcal{J}_{\text{GRPO}}(\theta)
&= \mathbb{E}\Bigg[
\frac{1}{G}\sum_{i=1}^{G}\frac{1}{|o_i|}\sum_{t=1}^{|o_i|}
\Bigg(
\min\Big(
i_t(\theta)\hat{A}_{i,t},\;
\operatorname{clip}\big(i_t(\theta), \\[-3pt]
&\qquad\qquad
1-\varepsilon,\,1+\varepsilon\big)\hat{A}_{i,t}
\Big)
-\beta\,D_{\mathrm{KL}}\big[\pi_\theta \,\|\, \pi_{\text{ref}}\big]
\Bigg)
\Bigg],
\end{aligned}
\label{eq:grpo}
\end{equation}
}

{GRPO} \citep{shao2024deepseekmath} is a scalable policy gradient method that eliminates the need for a learned value function with group-based relative advantages obtained by ``ranking'' multiple samples per prompt. This critic-free, low-variance update scales well to long sequences where value estimation could be impractical. Unlike PPO \citep{schulman2017proximal}, which relies on a critic, GRPO uses reward-standardized advantages together with PPO-style clipping and a KL penalty to a frozen reference policy, yielding efficient optimization.

For each prompt \(q\), the policy \(\pi_\theta\) generates a group of \(G\) candidate outputs \(\{y_1, y_2, \dots, y_G\}\). A composite reward is computed for each output using our multi-objective reward function (Section~\ref{subsec:reward}), and rewards are standardized within the group to compute the advantage \(\hat{A}_i\). Specifically:
\begin{equation}
\hat{A}_i \;=\; \frac{r_i - \bar{r}}{\sigma_r},
\label{eq:advantage}
\end{equation}
where \(r_i\) is the reward for \(y_i\), \(\bar{r}\) and \(\sigma_r\) are the group mean and standard deviation of rewards. Intuitively, outputs above the group average receive positive advantages, while those below receive negative ones.

% uses rank-based advantages together with PPO-style clipping and a KL penalty to a frozen reference policy, yielding efficient optimization. 

% For each prompt \(q\), the policy \(\pi_\theta\) generates a group of \(G\) candidate outputs \(\{y_1, y_2, \dots, y_G\}\). A composite reward is computed for each output using our multi-objective reward function (Section~\ref{subsec:reward}), and rewards are normalized within the group to compute the advantage \(\hat{A}_i\). Specifically:
% \begin{equation}
% \hat{A}_i = \frac{\text{rank}(y_i)}{G} - \bar{r},
% \label{eq:advantage}
% \end{equation}
% where \(\text{rank}(y_i)\) denotes the \emph{normalized} rank of response \(y_i\) in the group (higher is better, mapped to \([0,1]\)), and \(\bar{r}\) is the mean normalized rank serving as a baseline. Intuitively, outputs ranked higher than average receive positive advantages, while those below average receive negative ones, an approach that is more robust than raw scalar rewards for subjective measures like chart readability. 

%\joty{Why are you using rank in stead of reward directly? This needs better justification. What happens if all members in the group get 0 or 1?}
%\textcolor{red}{We do not use ranks. Our implementation follows GRPO in TRL where advantages are computed from within‑group standardized rewards: I updated the text. }  

The complete GRPO objective, which combines ratio clipping for stable policy updates and a KL regularization term to prevent excessive divergence from the reference policy, is shown in Equation~\ref{eq:grpo}, where $i_t(\theta) = \frac{\pi_\theta(y_{i,t} \mid q, y_{i,<t})}{\pi_{\text{old}}(y_{i,t} \mid q, y_{i,<t})}$ is the importance sampling ratio between the current (trainable) policy and the old frozen policy (previous iterate) for the $t$-th token in $y_i$. $\hat{A}_{i,t}$ is the within-group, normalized advantage as per Eq. \ref{eq:advantage} (every token in $y_i$ gets the same advantage); $\pi_{\text{ref}}$ is the fixed reference policy used in the KL term. Finally, $\varepsilon$ sets the PPO‑style clipping bound on the importance ratio $i_t(\theta)$ to stabilize updates, and $\beta$ weights the KL penalty $D_{\mathrm{KL}}[\pi_\theta|\pi_{\text{ref}}]$ to limit divergence from the reference policy.

% $\varepsilon$ and $\beta$ are two hyperparameters that \joty{.....say their role..}

%$\pi_{\text{old}}$ is the frozen behavior policy (previous iterate) used in the denominator; 

% The complete GRPO objective combining ratio clipping for stable policy updates and a KL regularization term to prevent excessive divergence from the reference policy is shown in Equation ~\ref{fig:grpo_compact}. For compactness, we denote by \(\mathcal{T}_i\) the per-token average over output \(o_i\), where \(\mathcal{T}_i := \tfrac{1}{|o_i|}\sum_{t=1}^{|o_i|}\). 

%\joty{I'd rather $i_t(\theta) = \frac{\pi_\theta\!\bigl(o_{i,t}\,|\,q,o_{i,<t}\bigr)}     {\pi_{\text{old}}\!\bigl(o_{i,t}\,|\,q,o_{i,<t}\bigr)}$ to simplify Eq. 3, $\mathcal{T}_i := \tfrac{1}{|o_i|}\sum_{t=1}^{|o_i|}$ does not take much space and it is important to show it in the Eq. Also mention what is $\pi_{old}$ and $\epsilon$.}
%\textcolor{red}{Updated Equation 3}

% GRPO offers several benefits over PPO: (i) it avoids training a separate critic, (ii) scales effectively to long sequences, and (iii) aligns naturally with multimodal reward models where explicit value estimation is infeasible. These properties make GRPO particularly suitable for our RL-Text2Vis framework, where the goal is to optimize structured, multi-step reasoning across text, code, and visualization.

\subsection{Multi-Objective Reward Design}
\label{subsec:reward}

A key challenge in Text2Vis generation lies in jointly optimizing correctness across text, code, and visualization dimensions. To address this, we adopt a two-stage reward system that ensures structural validity while providing fine-grained feedback for quality improvement.

\noindent\textbf{Stage 1: Format Reward.}  
The format reward enforces compliance with the required output schema, where the model must return a valid JSON object containing two fields: \texttt{answer} (a short textual response) and \texttt{code} (a runnable Python visualization script ending with \texttt{plt.show()}). Responses failing this structural constraint receive a reward of zero and are excluded from subsequent optimization. This step prevents policy drift toward invalid generations.

\noindent\textbf{Stage 2: Composite Reward.}  
For responses passing the format check, we compute a composite reward integrating three complementary signals:

\begin{enumerate}[leftmargin=14.5pt,labelsep=0.5em,itemsep=0pt,parsep=0pt,topsep=0pt]
    \item \textbf{Textual Correctness (\(R_{\text{text}}\))}:  
    We use an LLM-based evaluator to assess semantic alignment between the generated answer and the ground truth. The score ranges from 0 to 1, with partial credit for near-synonyms or numerically close values.
    
    \item \textbf{Code Reward (\(R_{\text{code}}\))}:  
    This combines two aspects with equal weight: (i) \emph{Executability}, verified by running the generated code in a sandboxed environment, and (ii) \emph{Intent Match}, scored by an LLM comparing generated code with reference query and code. We compute \(R_{\text{code}}=\tfrac{1}{2} I_{\text{exec}}+\tfrac{1}{2} I_{\text{intent}}\) with \(I_{\text{exec}}, I_{\text{intent}} \in \{0,1\}\), so \(R_{\text{code}} \in \{0, 0.5, 1\}\). 
    
    % \joty{give range and how you combine-updated} 
    % Non-executable code receives zero reward for this component.
    
    \item \textbf{Visualization Quality (\(R_{\text{vis}}\))}:  
    We leverage a VLM to evaluate the resulting chart on two dimensions: readability (layout clarity, label quality) and correctness (faithfulness to query and data). Both scores are normalized to [0,1] and averaged. 
\end{enumerate}
The final composite reward is expressed as a weighted sum:
\begin{equation}
R = \alpha R_{\text{text}} + \beta R_{\text{code}} + \gamma R_{\text{vis}}, \quad \text{with } \alpha+\beta+\gamma=1,
\tag{4}
\label{eq:composite-reward}
\end{equation}
where \(\alpha,\beta,\gamma\) control the relative emphasis on textual, code, and visual quality. The weighting strategy is a practical tuning step to stabilize training.
We performed a small grid search over \(\alpha \in \{0.3,0.4,0.5,0.6\}\), \(\beta \in \{0.2,0.25,0.3\}\), and \(\gamma = 1-\alpha-\beta \ge 0.1\),
and selected \((\alpha,\beta,\gamma) = (0.50,0.25,0.25)\) as the \textit{best trade-off} setting: it improved code executability and visualization quality without degrading textual correctness.
This choice maximized the overall validation metrics, indicating balanced performance across textual, code, and visual dimensions.

% \vspace{-2mm}

\subsection{Implementation Details}
We trained two open-source models, Qwen2.5-7B-Instruct and Qwen2.5-14B-Instruct, using the proposed RL-Text2Vis framework on the Text2Vis benchmark \cite{text2vis2025}. The Text2Vis dataset comprises two splits: \textit{test1} with 1,749 samples used for RL optimization, and \textit{test2} with 236 stratified samples used for final evaluation.

GRPO was run with a group size of \(G=8\) (completions per prompt) with within-group, reward-based advantage normalization. With a per-device batch of 8 prompts on 6 GPUs and gradient accumulation of 8, the rollout batch is 48 groups per micro-step and the effective batch is 384 groups per optimizer update. Rewards were computed using the two-stage scheme in Section~\ref{subsec:reward}. Specifically, the text-reward judge was Qwen2.5-7B for the 7B policy and Qwen2.5-14B for the 14B policy; the visual-reward judge was Qwen2.5-VL-7B and Qwen2.5-VL-32B, respectively. These judge models are used in an off-the-shelf manner, prompted without any fine-tuning. To mitigate evaluator bias, we conducted a stratified 200-sample comparison of rewards from Qwen2.5-7B, Qwen2.5-VL-7B, and GPT-4o, and observed strong cross-judge agreement (Pearson $r = 0.85$–$0.97$) (see Appendix~\ref{app:impl-details} for full details). 

For our selection of base models, we excluded coding models whose performance on Text2Vis were substantially worse, as this benchmark requires not only code generation but also reasoning over flattened tabular data and producing semantically aligned textual answers. Among open-source instruct models, Qwen2.5 achieved the strongest zero-shot performance on Text2Vis, motivating its selection as the primary backbone. To verify framework generality, we also trained a Llama-3.1-8B \cite{grattafiori2024llama} model under the same GRPO setup, which yielded comparable results and confirmed the framework’s model-agnostic applicability \ref{subsec:scaling}.

\section{Experiments} 
\subsection{Benchmarks}

We evaluate RL-Text2Vis on four benchmarks covering in-domain and out-of-domain settings:

\noindent \textbf{ (i) Text2Vis (In-Domain):} This benchmark is designed for natural language to visualization generation and includes natural language queries, tabular datasets, and reference outputs (answer and visualization code). Following the official evaluation split, \emph{test2} (236 stratified samples) is used strictly as a held‑out evaluation set; no hyperparameters are tuned on \emph{test2}. 

% We follow the benchmark’s original split names: the split labeled \emph{test1} (1,749 samples) is used \emph{only} for RL optimization, while \emph{test2} (236 stratified samples) is held out strictly for final evaluation; no hyperparameters are tuned on \emph{test2}. 

\noindent \textbf{(ii) VisEval (Out-of-Domain):} A benchmark with 2,524 queries across 146 databases, designed to evaluate visualization systems on validity, legality, and readability \cite{chen2024viseval}.  It features ambiguous mappings, challenging layouts, and diverse schemas. It does not provide ground-truth textual answers, so we report only code executability and visual quality (readability, correctness).  

\noindent \textbf{(iii) NVBench (Out-of-Domain):} A large-scale dataset with 7,247 visualization generation tasks across 105 domains and 153 datasets \cite{luo2021nvbench}. It supports evaluation on complex analytical queries and diverse chart types. Like VisEval, it does not include ground-truth textual answers, so Answer Correctness is omitted.

\noindent \textbf{(iv) PandasPlotBench (Out-of-Domain):} A benchmark of 175 data points designed to assess a model’s ability to generate executable and visually accurate plotting code from Pandas DataFrame \cite{galimzyanov2025drawing}.

% Across these three benchmarks, we evaluate Answer Correctness where available (Text2Vis), and Code Executability plus visual quality (readability and correctness) under both in-domain and out-of-domain conditions.   
\vspace{-2mm}
\subsection{Baselines}
We compare RL-Text2Vis against a comprehensive set of baselines covering closed-source and open-source LLMs, code-specialized models, and supervised fine-tuned variants. Among proprietary systems, GPT-4o \cite{openai2024gpt4technicalreport} and Gemini 1.5 Flash and 2.0 Flash \cite{geminiteam2024gemini15unlockingmultimodal} serve as state-of-the-art commercial models and strong zero-shot baselines. For open-source alternatives, we include LLaMA-3.1-8B, CodeLLaMA (7B and 13B) \cite{roziere2023code}, and Mistral-7B \cite{jiang2023mistral}. We also evaluate Qwen2.5 models \cite{yang2024qwen2} in both general-purpose and code-optimized (Coder-7B) variants under zero-shot settings to assess raw capability.  In addition, we include supervised fine-tuned baselines trained on the Text2Vis \textit{test1} split (5 epochs): Qwen2.5-7B SFT and Qwen2.5-14B SFT. We exclude DPO because it requires preference data, which is not readily available for Text2Vis. In our preliminary experiments, we also found GRPO to be more efficient—since it does not require a value model—and more effective than PPO. Therefore, we adopt GRPO as our RL algorithm. All systems (zero-shot baselines, the SFT model, and our RL-Text2Vis models) are decoded with temperature $=0.7$, top-$p$ $=0.9$, and a 2{,}048-token limit to ensure comparability.

%\textcolor{red}{We exclude PPO, DPO, and the weighted-reward setting as standalone baselines since they require preference data or pre-trained reward models, which are unavailable for Text2Vis; the weighted configuration is used solely for tuning rather than as a distinct comparison model.} 

% During RL training, we vary the number of sampled generations per input (e.g., 4 vs.\ 8) to study the effect of ranking-based updates. This setup ensures fair comparison across prompting, supervised fine-tuning, and reinforcement learning strategies.

% \textcolor{red}{As our primary contribution lies in a multi-objective post execution RL framework, we focus on comparisons with zero-shot, supervised fine-tuning (SFT), and single-reward RL baselines without post-execution feedback, which together represent the strongest non-RL and partial-RL references for Text2Vis. The weighted-reward variant is included only as part of hyperparameter tuning rather than a separate contribution. We exclude PPO and DPO baselines since both require explicit preference data or pre-trained reward models, neither of which are available in Text2Vis, where supervision consists solely of query–answer–code triplets and post-execution feedback.}

\vspace{-2mm}
\subsection{Evaluation Metrics}
\vspace{-1mm}
We follow the evaluation protocol introduced in \cite{text2vis2025} to assess text-to-visualization generation across four dimensions: (i) \textbf{Answer Correctness}, which measures whether the generated textual response matches the ground truth; (ii) \textbf{Code Executability}, which checks if the visualization code runs without errors; (iii) \textbf{Chart Readability}, which evaluates layout clarity and labeling quality; and (iv) \textbf{Chart Correctness}, which verifies whether the visualization accurately reflects the intended analytical insight. Answer correctness and code executability are treated as binary metrics, while chart readability and correctness are rated on a 1–5 scale. A sample is considered a pass if the code executes successfully, the answer matches the ground truth, and both the readability and chart correctness scores are at least 3.5. We use GPT-4o as the primary evaluator, as it has shown strong agreement with human ratings for both textual and visual assessments \cite{ chen2024viseval,gu2024survey}. To reduce evaluation bias, we perform manual evaluation on the official evaluation set (236 samples) across RL-Text2Vis (7B, 14B), Qwen2.5 (7B, 14B) zero-shot counterparts, GPT-4o, and the Qwen2.5 (7B, 14B) SFT baselines. See evaluation guidelines and metric definitions in Appendix Table~\ref{tab:result_evaluation}.

\begin{table*}[h!]
    \vspace{-2mm}
    \vskip -.5ex
\tiny
\centering
\caption{Performance on Text2Vis. Readability and correctness are rated on a 1–5 scale. Final Pass Rate (\%) reflects combined criteria.
% \enamul{bold best results}
}
\label{tab:text2vis_dual_finalrate}
\resizebox{1\columnwidth}{!}{%
\begin{tabular}{>{\raggedright\arraybackslash}p{3.5cm}|c|c|c|c|c}
\hline
\textbf{Model} &
\makecell{\textbf{Code Exec.}\\\textbf{Success (\%)}} &
\makecell{\textbf{Answer}\\\textbf{Match (\%)}} &
\makecell{\textbf{Visual Clarity}\\\textbf{Readability}} &
\makecell{\textbf{Chart}\\\textbf{Correctness}} &
\textbf{Final Pass Rate (\%)} \\
\hline
\rowcolor[HTML]{E5F1FB}GPT-4o (Zero-Shot)  &87   &\textbf{39}   &3.32   & 3.30  & \textbf{30}  \\
\rowcolor[HTML]{E5F1FB}Gemini-1.5-Flash (Zero-Shot)      & 82  & 32  & 3.26  & 2.95  & 16  \\
\rowcolor[HTML]{E5F1FB}Gemini-2.0-Flash (Zero-Shot)      & 90  & 35  & 3.73  & 3.66  &26   \\
\rowcolor[HTML]{E5F1FB}Mistral-7B (Zero-Shot)           & 42  & 22  &1.57   &1.39   & 7  \\

\rowcolor[HTML]{E5F1FB}Llama-3.1-8B (Zero-Shot)           & 70  & 25  &1.81   &1.62   & 8  \\
\rowcolor[HTML]{E5F1FB}CodeLlama-7B (Zero-Shot)           & 62  & 11  &2.34   &1.50   & 2  \\
\rowcolor[HTML]{E5F1FB}CodeLlama-13B (Zero-Shot)           & 54  & 16  &1.81   &1.45   & 4  \\
\rowcolor[HTML]{E5F1FB}Qwen-2.5-Coder-7B (Zero-Shot)           & 32  & 24  &1.37   &1.23   & 4  \\

% \rowcolor[HTML]{DFF0D8}Qwen-2.5-3B (Zero-Shot)           & 67  & 13  &1.94   &1.63   & 3  \\

\rowcolor[HTML]{DFF0D8}Qwen-2.5-7B (Zero-Shot)           & 77  & 27  &2.81   &2.69   & 13  \\

\rowcolor[HTML]{DFF0D8}Qwen-2.5-14B (Zero-Shot)          & 78  &29   & 3.12  & 2.94  & 14  \\

% \rowcolor[HTML]{FFF3CD}Qwen-2.5-3B (SFT)                 &   &  &   &   &   \\

\rowcolor[HTML]{FFF3CD}Qwen-2.5-7B (SFT)                 & 85  & 30 &3.34   &3.32   & 16  \\

\rowcolor[HTML]{FFF3CD}Qwen-2.5-14B (SFT)                 & 87  & 36 &3.42   &3.28   & 18  \\

% \rowcolor[HTML]{F2DEDE}\textbf{RL-Text2Vis-3B} (Gen-8)            &   &   &   &   &   \\

\rowcolor[HTML]{F2DEDE}\textbf{RL-Text2Vis-7B} (Gen-8)            & 91  & 31  &3.84   & 3.86  & 22  \\

\rowcolor[HTML]{F2DEDE}\textbf{RL-Text2Vis-14B} (Gen-8)           & \textbf{97}  & 35  & \textbf{4.10}  & \textbf{4.03}  & 29   \\
\hline
\end{tabular}}
    \vspace{-2mm}
    \vskip -.5ex
\end{table*}

% \vspace{-2mm}

\section{Results}

% \vspace{-2mm}

\subsection{In-Domain Results}  
Table~\ref{tab:text2vis_dual_finalrate} presents in-domain results on the Text2Vis benchmark. RL-Text2Vis substantially outperforms all open-source baselines across every metric. For example, the RL-Text2Vis 14B model improves chart readability from 3.12 to 4.10 and chart correctness from 2.94 to 4.03 compared to zero-shot Qwen2.5-14B.
%, reflecting a significant improvement in visualization quality. 
Similarly, code execution success rises from 78\% to 97\%, and answer match improves from 29\% to 35\%, demonstrating the benefit of multimodal reward optimization. 
% \textcolor{red}{We observe smaller gains on Answer Match relative to clarity and correctness, primarily because many queries in Text2Vis involve multi-step reasoning, multi-valued outputs, or are inherently unanswerable. Scaling from 7B to 14B yields modest improvements, indicating that larger models better capture semantic reasoning required for textual accuracy.}
In contrast, supervised fine-tuning offers only marginal gains over zero-shot models, highlighting the limitations of standard instruction tuning in visualization tasks. The RL-Text2Vis 7B model also shows large improvements, achieving a chart readability score of 3.84 compared to 2.81 and chart correctness of 3.86 compared to 2.69 for its zero-shot counterpart.

Notably, RL-Text2Vis closes the gap with proprietary models and even surpasses them on key visualization metrics. While the state-of-the-art GPT-4o model achieves a comparable final pass rate, RL-Text2Vis (7B and 14B) outperforms it in chart readability, visual clarity, and correctness. For example, the RL-Text2Vis 14B model achieves a chart readability score of 4.10 compared to 3.32 and chart correctness of 4.03 compared to 3.30, showing a clear margin. These results demonstrate that RL with multimodal feedback enables open-source models to match or even exceed closed-source systems in visualization quality.
%quality-sensitive visual dimensions.

% \vspace{-2mm}

\subsection{Out-of-Domain Generalization}

Table~\ref{tab:ood_dual_finalrate} reports out-of-domain performance on VIS-Eval and NVBench, where models are trained only on the Text2Vis \textit{test1} set and evaluated without further adaptation. RL-Text2Vis demonstrates strong generalization across both benchmarks, substantially outperforming zero-shot Qwen baselines. On VIS-Eval, which is particularly challenging due to multi-table schemas, RL-Text2Vis-7B improves chart readability from 1.50 to 2.50 and chart correctness from 0.69 to 1.37, along with a notable gain in code execution success (from 57\% to 72\%).

% These results confirm that optimizing for semantic correctness, code validity, and visual quality during RL training yields models that generalize beyond the training domain, even in settings with substantially different data structures and visualization requirements. 

% These gains indicate that RL with multimodal rewards enhances robustness in handling schema diversity and complex visualization logic.

% Table~\ref{tab:ood_dual_finalrate} summarizes results on VIS-Eval and NVBench. Since these datasets lack answer annotations, we evaluate only on Code Execution Success, Visual Clarity, and Chart Correctness.

\begin{table}[t]
    \vspace{-2mm}
    \vskip -.5ex
\centering

\caption{Out-of-domain results on VIS-Eval, NVBench \& PandasPlotBench (readability, correctness: 1–5)
%Readability and correctness rated on a 1–5 scale.
}

% \enamul{bold best results}

\label{tab:ood_dual_finalrate}
\resizebox{1\columnwidth}{!}{%
\begin{tabular}{>{\raggedright\arraybackslash}p{4.2cm}|c|c|c}
\hline
\textbf{Model} &
\makecell{\textbf{Code Exec.}\\\textbf{Success (\%)}} &
\makecell{\textbf{Visual Clarity}\\\textbf{Readability}} &
\makecell{\textbf{Chart}\\\textbf{Correctness}} \\
\hline
\multicolumn{4}{l}{\textbf{VIS-Eval}} \\
\rowcolor[HTML]{E5F1FB}Qwen-2.5-7B (Zero-Shot)        & 57  & 1.50  & 0.69  \\
\rowcolor[HTML]{E5F1FB}Qwen-2.5-7B (SFT)        & 64  & 1.86  & 0.87  \\
\rowcolor[HTML]{F2DEDE}\textbf{RL-Text2Vis-7B} (Gen-8)         & 72  & 2.50  &1.37   \\
\rowcolor[HTML]{F2DEDE}\textbf{RL-Text2Vis-14B} (Gen-8)        & \textbf{74}  & \textbf{2.58}  & \textbf{1.48}  \\
\hline
\multicolumn{4}{l}{\textbf{NVBench}} \\
\rowcolor[HTML]{E5F1FB}Qwen-2.5-7B (Zero-Shot)        &75   & 2.64  & 2.34  \\
\rowcolor[HTML]{E5F1FB}Qwen-2.5-7B (SFT)        & 82  & 3.07  & 2.79  \\
\rowcolor[HTML]{F2DEDE}\textbf{RL-Text2Vis-7B} (Gen-8)         & 93  & 3.47  & 3.28  \\
\rowcolor[HTML]{F2DEDE}\textbf{RL-Text2Vis-14B} (Gen-8)        & \textbf{96}  & \textbf{3.95}  & \textbf{3.59}  \\
\hline

\multicolumn{4}{l}{\textbf{PandasPlotBench}} \\
\rowcolor[HTML]{E5F1FB}Qwen-2.5-7B (Zero-Shot)        &65   & 2.42  & 2.49  \\
\rowcolor[HTML]{F2DEDE}\textbf{RL-Text2Vis-7B} (Gen-8)         & 75  & 3.32  & 3.37  \\
\rowcolor[HTML]{F2DEDE}\textbf{RL-Text2Vis-14B} (Gen-8)        & \textbf{79}  & \textbf{3.65}  & \textbf{3.63}  \\
\hline

\end{tabular}}
    \vspace{-2mm}
    \vskip -.5ex
\end{table}

On NVBench, which includes large-scale cross-domain tasks, RL-Text2Vis achieves even greater improvements.  Code execution success improves from 75\% to 93\%, while chart readability rises from 2.64 to 3.47 and chart correctness from 2.34 to 3.28. On PandasPlotBench, which evaluates executable plotting from Pandas DataFrames, RL-Text2Vis maintains strong generalization, further validating its robustness across diverse visualization domains. These results show that optimizing semantic correctness, code validity, and visual quality during RL training improves generalization beyond the training domain, including to datasets with different schemas and chart types.

% On NVBench, which includes large-scale cross-domain tasks, RL-Text2Vis achieves even greater improvements. The 14B variant raises readability from 2.64 to 3.95 and correctness from 2.34 to 3.59, while execution success jumps from 75\% to 96\%, clearly outperforming zero-shot baselines. These results confirm that optimizing for semantic correctness, code validity, and visual quality during RL training yields models that generalize beyond the training domain, even in settings with substantially different data structures and visualization requirements. 

% \vspace{-2mm}

\subsection{Human Evaluation}
While GPT-4o serves as our primary evaluator due to its strong alignment with human judgments \cite{gu2024survey,text2vis2025}, we also conducted a manual study to ensure robustness. Two annotators independently assessed the full stratified official evaluation set of the Text2Vis benchmark (236 samples) using a structured rubric (Table~\ref{tab:result_evaluation}) covering label clarity, color use, font size, and overall visual appeal. They evaluated outputs from all major baselines—RL-Text2Vis (7B/14B), Qwen2.5 (zero-shot/SFT), and GPT-4o—while blinded to automated scores. Agreement between human and automated judgments was consistently strong, with Pearson correlation coefficients of $r=0.91$ for Answer Match, $r=0.88$ for Clarity \& Readability, and $r=0.88$ for Chart Correctness. In addition, code executability is measured objectively by verifying whether the generated Matplotlib code executes without errors, making this dimension independent of both human and automated judgments.  These results confirm that our reported improvements are reliably supported by both human and automated evaluation.

\begin{table}[!h]
    \vspace{-2mm}
    \vskip -.5ex
\centering
\caption{{Ablation on reward components (Qwen2.5-7B) on Text2Vis.}}
\label{tab:ablation_rewards_full}
\setlength{\tabcolsep}{6pt}  
\renewcommand{\arraystretch}{1.1}   
\Large                                
\begin{adjustbox}{max width=\columnwidth}
% \resizebox{1\columnwidth}{!}{%
\begin{tabular}{>{\raggedright\arraybackslash}p{5.5cm}|c|c|c|c|c}
\hline
\textbf{Configuration} &
\makecell{\textbf{Code Exec.}\\\textbf{Success (\%)}} &
\makecell{\textbf{Answer}\\\textbf{Match (\%)}} &
\makecell{\textbf{Visual Clarity}\\\textbf{Readability}} &
\makecell{\textbf{Chart}\\\textbf{Correctness}} &
\makecell{\textbf{Final Pass}\\\textbf{Rate (\%)}}\\
\hline
% \rowcolor[HTML]{E5F1FB}Format Reward Only         &78   & 25   &3.02   &2.88 &13  \\
% \rowcolor[HTML]{E5F1FB}Answer Reward Only         & 76  &30   &2.72   &2.63   & 11  \\
% \rowcolor[HTML]{E5F1FB}Code Reward Only           &86 & 26  &   2.92   & 2.81 & 14  \\
% \rowcolor[HTML]{E5F1FB}Visualizaion Reward Only  &79   &26   &3.22   &3.05   &14   \\

% \rowcolor[HTML]{E5F1FB}Format \& Answer Reward  & 82  & 28   & 2.98  & 3.10 & 13  \\

% \rowcolor[HTML]{E5F1FB}Code \&Visual. Reward  & 87   & 25   &    3.38  & 3.25 & 17 \\

\rowcolor[HTML]{F2DEDE}Full (All Rewards)    & \textbf{91}  & \textbf{31}  & \textbf{3.84}  & \textbf{3.86}  & \textbf{22}  \\

\rowcolor[HTML]{E5F1FB}- Format Reward  & 87   & 25   &    3.38  & 3.25 & 17 \\

\rowcolor[HTML]{E5F1FB}- Answer Reward  & 89   & 24   &    3.79  & 3.82 & 19 \\ 

\rowcolor[HTML]{E5F1FB}- Code \& Visual Reward  & 82  & 28   & 2.98  & 3.10 & 13  \\

\rowcolor[HTML]{FFF8DC}Format Reward Only         &78   & 25   &3.02   &2.88 &13  \\
\rowcolor[HTML]{FFF8DC}Answer Reward Only         & 76  &30   &2.72   &2.63   & 11  \\
\rowcolor[HTML]{FFF8DC}Code Reward Only           &86 & 26  &   2.92   & 2.81 & 14  \\
\rowcolor[HTML]{FFF8DC}Visual Reward Only  &79   &26   &3.22   &3.05   &14   \\
\hline
\end{tabular}
\end{adjustbox}
    \vspace{-2mm}
    \vskip -.5ex
\end{table}

% \vspace{-2mm}

\subsection{Ablation Studies}

% \vspace{-2mm}

\paragraph{Effect of Individual Reward Signals.} 
We perform ablations on the RL-Text2Vis-7B model to analyze the contribution of each reward component and their combination. Table~\ref{tab:ablation_rewards_full} reports two complementary analyses: removal-based ablations, where each component is dropped from the full model, and single-component ablations, where each reward is used independently. The results clearly show that using the multimodal reward yields the highest overall performance, confirming the effectiveness of jointly optimizing textual accuracy, code validity, and visualization quality.

% We perform ablations on the Qwen2.5-7B model to analyze the contribution of individual reward components and their combination. The four components Format, Answer, Code, and Visualization are described in Section~\ref{subsec:reward}. Table~\ref{tab:ablation_rewards_full} reports results when each component is applied independently and when all are combined in the full multi-objective setting. The results clearly show that using the multimodal reward yields the highest overall performance, confirming the effectiveness of jointly optimizing textual accuracy, code validity, and visualization quality.

% \vspace{-4mm}

\paragraph{Effect of Number of Sampled Completions.} 
To assess the effect of sample size during GRPO optimization, we trained RL-Text2Vis-7B model with 4 versus 8 completions per input. Table~\ref{tab:ablation_groupsize_full} shows that using more completions stabilizes ranking-based updates and yields consistent gains.
% Although our RL-Text2Vis models were trained with 8 completions, to analyze the impact of varying the number of sampled completions per input during GRPO optimization, we also trained a Qwen2.5‑7B model with 4 completions.
% Table~\ref{tab:ablation_groupsize_full} compares performance when sampling 4 versus 8 completions for ranking-based updates. Increasing the number of completions improves the stability of relative ranking, leading to higher-quality policy updates. This results in notable gains in visualization quality, with chart readability improving from 3.62 to 3.84 and chart correctness from 3.52 to 3.86, as well as an increase in Final Pass Rate from 17\% to 22\%.
These findings suggest that sampling more candidate outputs per input during preference optimization provides stronger learning signals, yielding more reliable and interpretable charts.

% \vspace{-2mm}

\subsection{Scaling and Cross-Architecture Analysis}
\label{subsec:scaling}
\vspace{-1mm}
To assess scalability and model generalizability, we trained RL-Text2Vis on both smaller (3B) and alternative (Llama-3.1-8B) architectures under the same GRPO setup. As shown in Table~\ref{tab:scaling}, applying RL consistently improved all metrics across scales and model families. For Qwen2.5-3B, RL-Text2Vis increased code executability from 67\% to 88\% and enhanced chart readability and correctness; however, textual answer accuracy remained unchanged, revealing that smaller models can learn syntactic and visual patterns but lack sufficient capacity for reasoning improvements. Similarly, on Llama-3.1-8B, RL-Text2Vis improved all metrics, demonstrating effective transfer beyond the Qwen architecture. These results confirm that our GRPO-based framework is architecture-agnostic and scales robustly across model families and sizes.

\begin{table}[t]
\centering
\caption{{Scaling and Cross-Architecture Results on Text2Vis.}}
\label{tab:scaling}

\setlength{\tabcolsep}{6pt}  
\renewcommand{\arraystretch}{1.2}   
\Large                                
\begin{adjustbox}{max width=\columnwidth}
\resizebox{1\columnwidth}{!}{%
\begin{tabular}{>{\raggedright\arraybackslash}p{6.25cm}|c|c|c|c|c}

\hline
\textbf{Configuration} &
\makecell{\textbf{Code Exec.}\\\textbf{Success (\%)}} &
\makecell{\textbf{Answer}\\\textbf{Match (\%)}} &
\makecell{\textbf{Visual Clarity}\\\textbf{Readability}} &
\makecell{\textbf{Chart}\\\textbf{Correctness}} &
% \textbf{Final Pass Rate (\%)} \\
\makecell{\textbf{Final}\\\textbf{Pass Rate (\%)}} \\
\hline
\rowcolor[HTML]{E5F1FB}Qwen2.5-3B (Zero-Shot)   &   67&13   & 1.94  &1.63   &  3 \\
\rowcolor[HTML]{F2DEDE}RL-Text2Vis-3B   &\textbf{88}  & \textbf{12}  &\textbf{2.23}   & \textbf{1.91}  & \textbf{4}  \\
\rowcolor[HTML]{E5F1FB}Llama-3.1-8B (Zero-Shot)   &70   & 25  & 1.81  & 1.62  &8 \\
\rowcolor[HTML]{F2DEDE}RL-Text2Vis-Llama-3.1-8B   &\textbf{87}  & \textbf{28}  &\textbf{2.91}   & \textbf{2.67}  & \textbf{15}  \\
\hline
\end{tabular}}
\end{adjustbox}
    \vspace{-2mm}
    \vskip -.5ex
\end{table}
\vspace{-2mm}
\section{Error Analysis}
\label{sec:error_analysis}

We manually evaluated all samples generated by the zero-shot Qwen2.5-14B baseline and our RL-Text2Vis-14B model to analyze failure patterns.  Common failures included \textit{syntax errors} (e.g., ``invalid syntax'' or missing commas), \textit{shape mismatches} (e.g., ``x and y must have same first dimension''), and \textit{value errors} such as incorrect CAGR or percentage calculations. Additional problems involved missing imports (e.g., \texttt{name 'combinations' is not defined}), logic inconsistencies (e.g., unterminated strings), and glyph-related warnings. Beyond code-level issues, many visualizations exhibited low readability, poor chart layout, missing axis labels, and weak alignment with query intent. 

Applying GRPO-based RL significantly mitigated these challenges by leveraging post-execution feedback in a multi-objective reward design. Figure~\ref{fig:error_analysis}  illustrates several representative improvements: (1) syntax and structural errors were eliminated; (2) readability and visual clarity improved through format and visualization rewards; (3) value and logic errors were corrected by enforcing semantic alignment; (4) missing labels were added; (5) charts became query-aligned and visually interpretable; and (6) type errors and invalid operations were resolved.

%% file: Appendix.tex
% \cleardoublepage
\appendix
\section{Appendices}
\label{app:Appendice}

\subsection{Reinforcement Learning for LLM Alignment}
\label{app:rl}

\textbf{Reinforcement Learning (RL)} provides a framework for optimizing sequential decision-making by learning policies that maximize expected cumulative rewards. Unlike supervised learning, which relies on explicit ground-truth labels, RL agents learn from feedback signals (rewards) derived from interactions with an environment. This paradigm has been successfully applied to robotics, games, and more recently to aligning large language models (LLMs) with human preferences.

\subsubsection{Reinforcement Learning from Human Feedback (RLHF)}
RLHF \citep{ouyang2022training} aligns LLMs with human preferences through three steps:  
(i) \textit{Supervised Fine-Tuning (SFT)} on curated demonstrations,  
(ii) training a \textit{reward model} from human comparisons, and  
(iii) optimizing the policy using a reinforcement learning algorithm, typically Proximal Policy Optimization (PPO).  
While highly effective, RLHF is computationally expensive.

\subsubsection{Proximal Policy Optimization (PPO)}
PPO \citep{schulman2017proximal} is the foundation of RLHF and provides stable policy updates via a clipped surrogate objective:

\begin{equation}
\small
\begin{aligned}
\mathcal{L}_{\text{PPO}} = \mathbb{E} \Bigg[ 
\min \Big( r_t(\theta)\hat{A}_t,\;  
\text{clip}\big(r_t(\theta), 1-\epsilon, 1+\epsilon\big)\hat{A}_t \Big)
\Bigg],
\end{aligned}
\end{equation}

where $r_t(\theta)$ is the probability ratio between new and old policies, and $\hat{A}_t$ is the advantage function. PPO employs a value function (critic) for advantage estimation and uses KL regularization for stable updates.

\subsubsection{Direct Preference Optimization (DPO)}
DPO eliminates the need for a reward model by optimizing the policy directly on human preference data. Given a preferred response $y^+$ and a dispreferred response $y^-$ for the same input $x$, DPO maximizes:

\begin{equation}
\small
\begin{aligned}
\mathcal{L}_{\text{DPO}} = 
-\log \sigma \Bigg( 
\beta \Big[ 
\log \frac{\pi_\theta(y^+|x)}{\pi_{\text{ref}}(y^+|x)} 
- \log \frac{\pi_\theta(y^-|x)}{\pi_{\text{ref}}(y^-|x)} 
\Big] \Bigg),
\end{aligned}
\end{equation}

where $\beta$ controls the sharpness of preference alignment. Unlike PPO, DPO skips the RL loop and directly uses preference pairs to guide updates.

\subsubsection{Group Relative Policy Optimization (GRPO)}

GRPO \citep{shao2024deepseekmath} extends PPO for settings where explicit value functions are impractical. Instead of using a critic, GRPO estimates advantages via \emph{within-group reward standardization}. For each input, the model generates a group of $G$ candidate outputs, assigns a scalar reward $r_i$ to each, and computes the normalized advantage:

\begin{equation}
\hat{A}_i = \frac{r_i - \bar{r}}{\sigma_r}, 
\end{equation}

where $\bar{r}$ and $\sigma_r$ are the group mean and standard deviation.

The GRPO objective follows a PPO-style clipped surrogate with KL regularization:

\begin{equation}
\small
\begin{aligned}
\mathcal{L}_{\text{GRPO}} = \mathbb{E} \Big[ 
\min \big( r_i(\theta)\hat{A}_i,\;
\text{clip}(r_i(\theta), 1-\epsilon, 1+\epsilon)\hat{A}_i \big) 
\Big] \\
- \beta D_{\text{KL}}\big(\pi_\theta \,\|\, \pi_{\text{ref}}\big).
\end{aligned}
\end{equation}

GRPO is well-suited for tasks like RL-Text2Vis, where key metrics—textual correctness, code executability, and visualization clarity—are only observable post-execution.

\subsubsection{Key Differences: DPO vs. GRPO}
Both methods optimize from preference signals, but DPO operates on pairwise preferences without an RL loop, while GRPO generalizes to groups and retains PPO-like stability. It is more suitable for tasks requiring structured outputs and multimodal optimization.

\subsection{Additional Implementation Details} 
\label{app:impl-details}

\subsubsection{In-loop evaluators}

To match real-world deployment and privacy constraints, we prioritized open-source evaluators for in-loop rewards. For the visual reward, we first compared Qwen2.5-VL-3B with LLaMA-3.2-3B VLM on a 100-sample set and found that Qwen achieved better agreement with human ratings on visualization quality. Based on this, we adopted Qwen2.5-VL-3B (and its larger variants) as the in-loop visual judge. For text rewards, we used Qwen2.5-7B for the 7B policy and Qwen2.5-14B for the 14B policy; for visual rewards, we used Qwen2.5-VL-7B and Qwen2.5-VL-32B, respectively. To mitigate evaluator bias, we conducted a stratified 200-sample comparison of rewards from Qwen2.5-7B, Qwen2.5-VL-7B, and GPT-4o, and observed strong cross-judge agreement (Pearson $r = 0.85$–$0.97$), indicating that the policy did not overfit to any single evaluator’s biases.

% We trained two open-source models, Qwen2.5-7B-Instruct and Qwen2.5-14B-Instruct, using the proposed RL-Text2Vis framework on the Text2Vis benchmark \cite{text2vis2025}. %Qwen2.5 is a state-of-the-art open-source model at these parameter scales and showed better zero-shot performance on the Text2Vis task. 
% The Text2Vis dataset consists of two splits: \textit{test1}, which contains 1,749 samples and is used for RL optimization, and \textit{test2}, which includes 236 stratified samples for final evaluation. GRPO was run with a group size of \(G=8\) \emph{completions per prompt}, using within-group reward-based advantage normalization. With a per-device batch size of 8 prompts on 6 GPUs and gradient accumulation of 8, this yields a rollout batch of 48 groups per micro-step and an effective batch of 384 groups per optimizer update. Rewards were computed using the two-stage scheme described in Section~\ref{subsec:reward}.  

\subsubsection{Hardware and optimization}

Training was conducted on high-performance hardware: Qwen2.5-7B-Instruct was fine-tuned on 4$\times$ NVIDIA A100 80\,GB GPUs, and Qwen2.5-14B-Instruct on 6$\times$ NVIDIA H100 80\,GB GPUs, both for two epochs. We used AdamW (learning rate $1 \times 10^{-5}$, weight decay $0.1$) with a cosine scheduler, KL penalty $\beta = 0.04$,
and gradient‑norm clipping $=0.1$. Gradient checkpointing and \texttt{bf16} mixed precision were enabled to reduce memory usage. These hyperparameters were selected via a small grid search, choosing the configuration that maximized  performance on a fixed 10\% development subset of Text2Vis \texttt{test1}. The full training process required approximately 25\,hours for the 7B model and 50\,hours for the 14B model.

\subsection{Error Analysis} 

\begin{figure*}[t!]
    \centering
    \includegraphics[width=\textwidth]{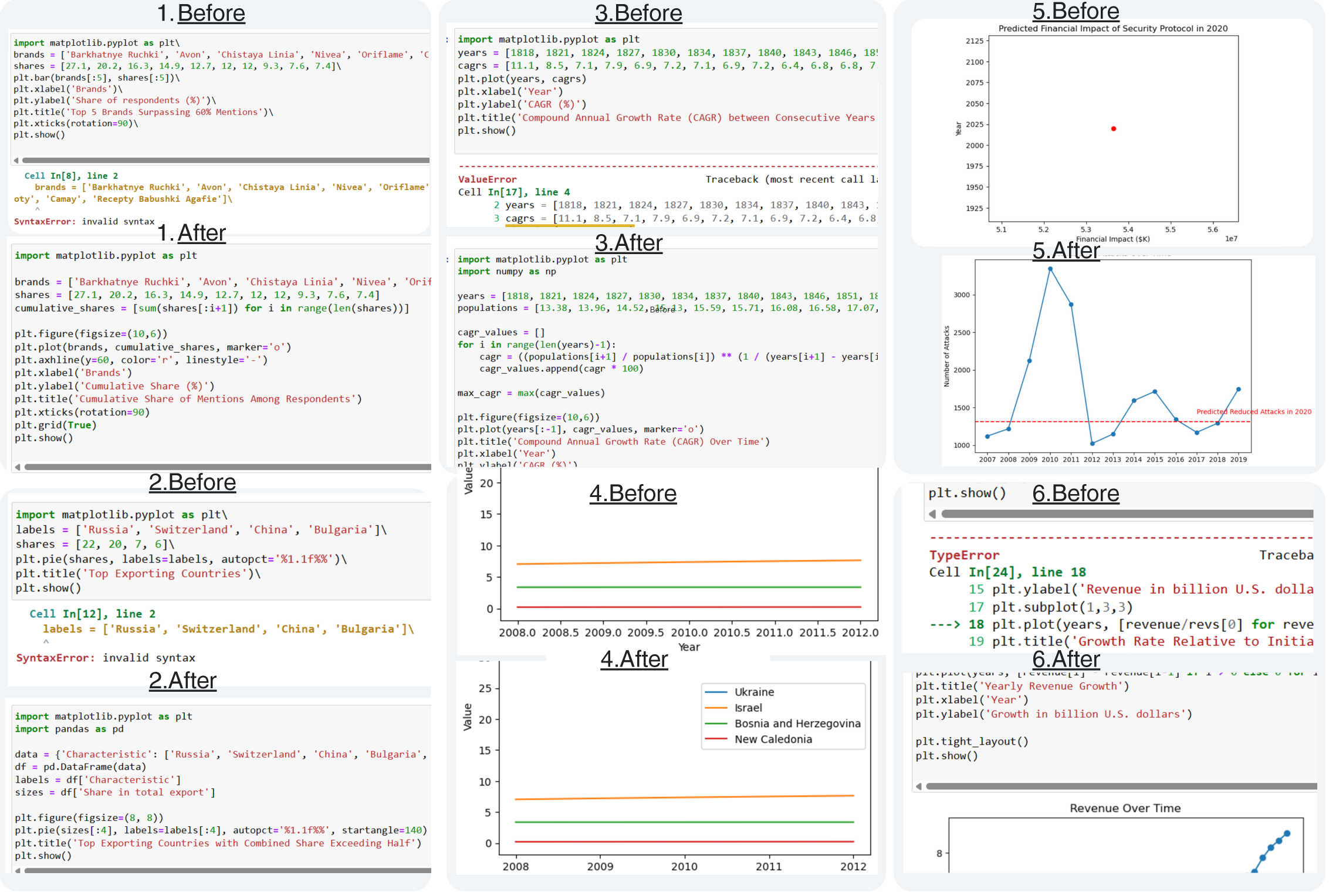}
    \caption{\textbf{Error analysis before and after GRPO.} GRPO significantly improves text-to-visualization generation by resolving errors such as syntax,  value errors, enhancing readability, visual quality, and alignment with the query.}    
    \label{fig:error_analysis}
\end{figure*}

Figure~\ref{fig:error_analysis} presents a qualitative comparison between the zero-shot Qwen2.5-14B model and the RL-Text2Vis-14B model. The analysis highlights that GRPO significantly improves visualization quality by reducing common issues such as non-executable code, misaligned charts, and low readability. Compared to baselines, our RL-trained models produce outputs that are more interpretable and semantically aligned with the query. 

\begin{table}[t]
\centering
\caption{Impact of GRPO group size on Qwen2.5-7B.}
\label{tab:ablation_groupsize_full}

\setlength{\tabcolsep}{5.25pt}  
\renewcommand{\arraystretch}{1}   
\Large                                
\begin{adjustbox}{max width=\columnwidth}
\resizebox{1\columnwidth}{!}{%
\begin{tabular}{>{\raggedright\arraybackslash}p{5.25cm}|c|c|c|c|c}
\hline
\textbf{Configuration} &
\makecell{\textbf{Code Exec.}\\\textbf{Success (\%)}} &
\makecell{\textbf{Answer}\\\textbf{Match (\%)}} &
\makecell{\textbf{Visual Clarity}\\\textbf{Readability}} &
\makecell{\textbf{Chart}\\\textbf{Correctness}} &
% \textbf{Final Pass Rate (\%)} \\
\makecell{\textbf{Final}\\\textbf{Pass Rate (\%)}} \\
\hline
\rowcolor[HTML]{E5F1FB}RL-Text2Vis-7B (Gen-4)   &87   &28   & 3.62  & 3.52  & 17  \\
\rowcolor[HTML]{F2DEDE}RL-Text2Vis-7B (Gen-8)   &\textbf{91}  & \textbf{31}  &\textbf{3.84}   & \textbf{3.86}  & \textbf{22}  \\
\hline
\end{tabular}}
\end{adjustbox}
\end{table}

\subsection{Training Dynamics}

We visualize sample training dynamics over 150 RL steps for RL-Text2Vis-14B. Each figure highlights different metrics tracked during GRPO training.

\begin{figure*}[t]
    \centering
    \includegraphics[width=\textwidth]{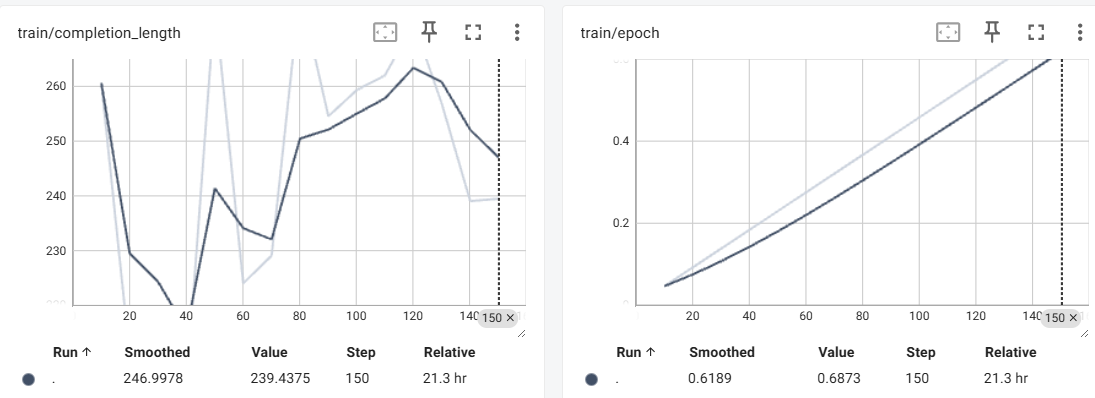}

    \caption{Completion length and epoch progression. The model stabilizes in output length while training steps progress linearly.}
    \label{fig:training_dynamics}
\end{figure*}

\begin{figure*}[t]
    \centering
    \includegraphics[width=\textwidth]{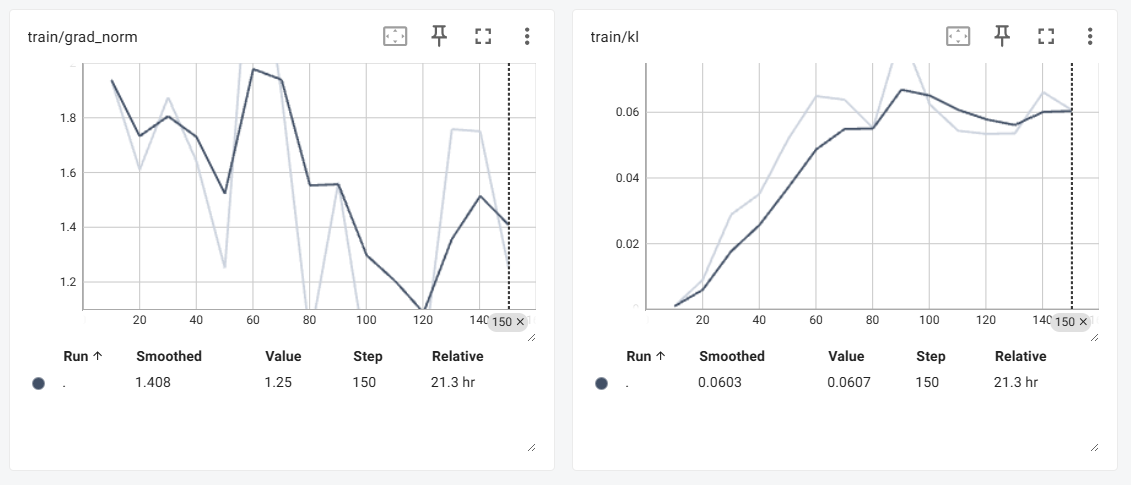}
     \caption{Gradient norm and KL divergence. Indicates optimization stability and policy deviation from the reference model.}
     \label{fig:training_dynamics_lr_loss}
\end{figure*}

\begin{figure*}[t]
    \centering
    \includegraphics[width=\textwidth]{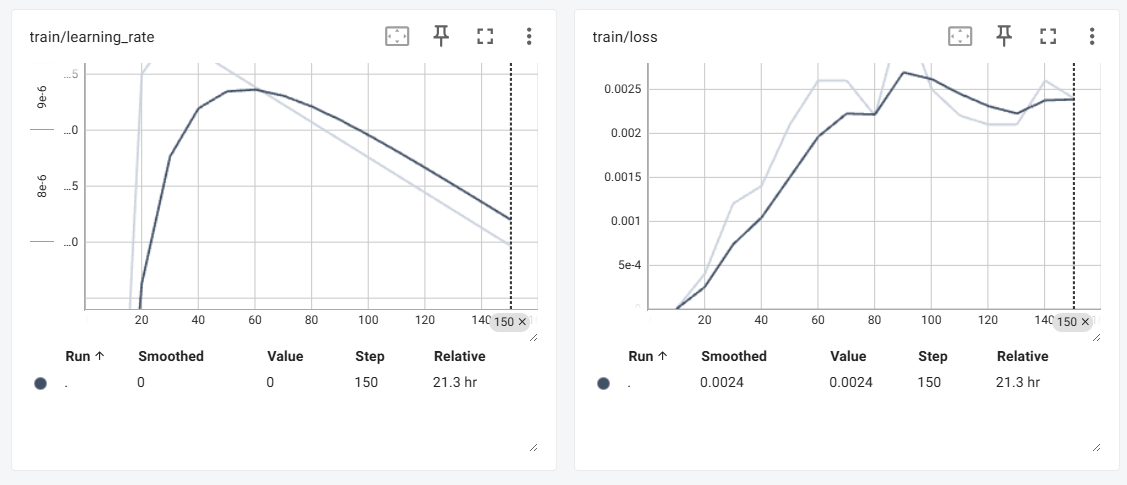}
    \caption{Learning rate schedule and training loss. Warm-up and decay patterns are followed, with loss trends influenced by reward-maximizing objectives.}
    \label{fig:training_dynamics_reward}
\end{figure*}

\begin{figure*}[t]
    \centering
    \includegraphics[width=\textwidth]{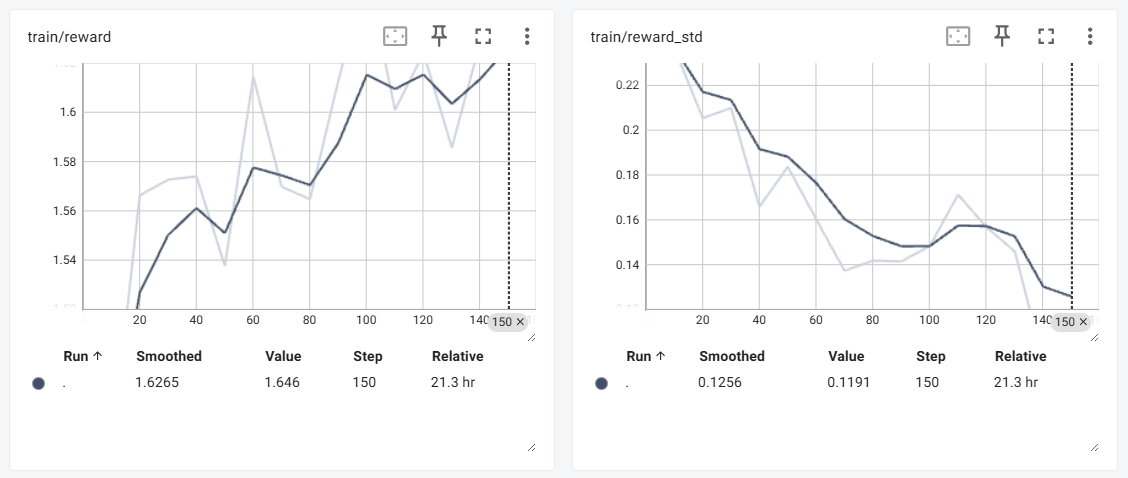}
    \caption{Average reward and reward standard deviation. Reflects growing reward consistency and reduced variance across outputs.}
    \label{fig:training_dynamics_reward}
\end{figure*}

\begin{figure*}[t]
    \centering
    \includegraphics[width=\textwidth]{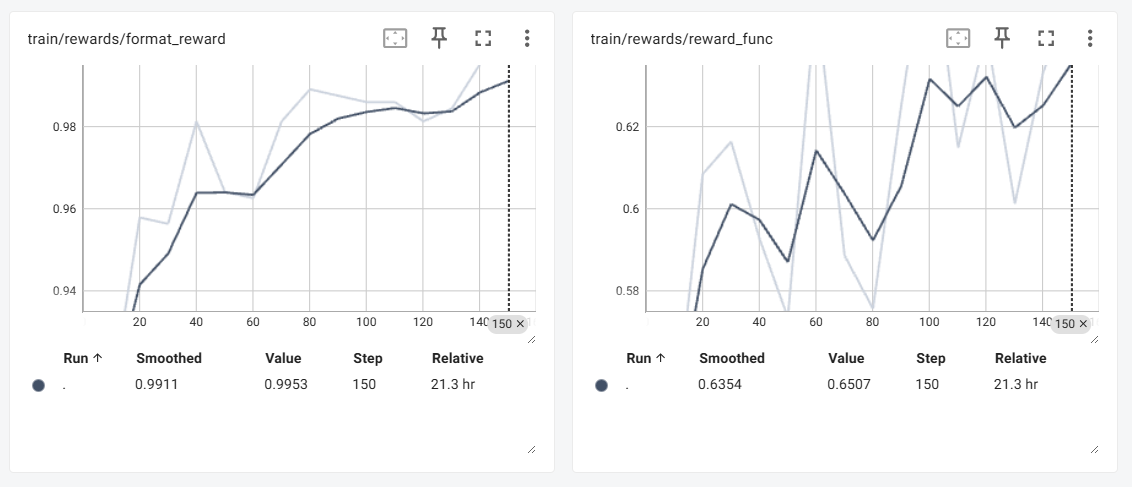}
    \caption{Format reward and Composite reward function score. Tracks alignment with structural and multimodal feedback objectives.}
    \label{fig:training_dynamics_rewards}
\end{figure*}

\subsection{Prompt Templates for Evaluation Metrics}
For all automated evaluations, we followed the official prompt templates provided in the Text2Vis benchmark \cite{text2vis2025}. These templates define instructions for assessing answer correctness, code executability, and visualization quality across readability and correctness dimensions.  

\begin{table*}[t]
  \centering
  \scriptsize
  \caption{Prompt Template for Evaluating Results Using the GPT-4.o Model \cite{text2vis2025} }
  \label{tab:result_evaluation}
  \rowcolors{2}{gray!10}{white}
  \begin{tabularx}{\textwidth}{>{\bfseries}l X}
    \toprule
    Category & Prompt Template \\
    \midrule
    \makecell[tl]{Evaluation}  & 
    \begin{minipage}[t]{\linewidth}
You are an evaluation expert responsible for assessing the accuracy of generated answers and the quality of visualizations. Given a structured \textbf{data table}, a user-generated question, a model-generated response, and an image-based visualization, your task is to validate the correctness of the response and evaluate the visualization quality.

\textbf{Input Data:}  
\begin{itemize}
    \item \textbf{Data Table}: \{\texttt{row['Table Data']}\}
    \item \textbf{Question}: \{\texttt{row['Generated Question']}\}
    \item \textbf{Generated Answer}: \{\texttt{row['Generated Answer']}\}
    \item \textbf{Ground Truth Answer}: \{\texttt{row['Answer']}\}
    \item \textbf{Generated Image}: \{\texttt{row['Generated image']}\}

\end{itemize}

\textbf{Task:}
\begin{enumerate}
    \item \textbf{Answer Matching}: Compare the generated answer with the ground truth using following evaluation  criteria.
    \item \textbf{Visualization Evaluation}: Score the visualization based on following evaluation  criteria.
\end{enumerate}

\textbf{Evaluation Criteria:}
\begin{enumerate}
    \item \textbf{Answer Matching (Binary: 1 or 0)}
    \begin{itemize} \tiny
        \item Match if numbers are close (e.g., "48.77" vs "48.73") or equivalent percentage formats (e.g., "100" vs "100%").
        \item Match if the ground truth appears within the generated response (e.g., "100" in "The result is 100").
        \item For long ground truth answer, match is considered as long as the core summary remains the same, even if the wording differs.
        \item Allow minor spelling variations or abbreviations (e.g., "Albenia" vs "Albania", "USA" vs "United States").
        \item No match if the meaning changes significantly (e.g., "Fragile" vs "Extreme fragility").
    \end{itemize}

    \item \textbf{Readability and Quality Score (0-5)}
    \begin{itemize} \tiny
        \item \textbf{Labels and Titles}: Are they clear, concise, and correctly positioned?
        \item \textbf{Layout Spacing}: Is the layout well-organized with no clutter?
        \item \textbf{Color Accessibility}: Are colors distinct and accessible (colorblind-friendly)?
        \item \textbf{Axis Scaling}: Are axes correctly labeled and proportional?
        \item \textbf{Chart Type Suitability}: Is the visualization appropriate for the data type (e.g., line chart for trends)?
        \item \textbf{Font and Legends}: Are fonts readable, and legends properly aligned?
        \item\textbf{Annotation Readability}: Are annotations (e.g., data labels, callouts) clear, well-placed, and non-overlapping?
    \end{itemize}

    \item \textbf{Chart Correctness Score (0-5)}
    \begin{itemize} \tiny
        \item \textbf{Query Alignment}: Does the visualization correctly address the question?
        \item \textbf{Data Integrity}: Are all data points accurately plotted?
        \item \textbf{Insight Representation}: Does the chart effectively communicate its key insights based on its type?
        \item \textbf{Handling Missing Data}: Is missing data presented appropriately without misleading distortion?
        \item \textbf{Complexity Handling}: For multi-step queries, is the visualization logically structured?
    \end{itemize}
\end{enumerate}

\begin{itemize}
    \item \textbf{5.0} – Excellent: Clear, accurate, and no issues.
    \item \textbf{4.5} – Very Good: Minor issues but does not impact understanding.
    \item \textbf{4.0} – Good: Small flaws like minor misalignments.
    \item \textbf{3.5} – Decent: Some readability/accuracy issues but still interpretable.
    \item \textbf{3.0} – Average: Noticeable problems that affect clarity or correctness.
    \item \textbf{2.5} – Below Average: Several issues that may lead to misinterpretation.
    \item \textbf{2.0} – Poor: Significant issues making the chart unclear.
    \item \textbf{1.5} – Very Poor: Major readability or correctness flaws.
    \item \textbf{1.0} – Unusable: Completely unclear or misleading.
    \item \textbf{0.0} – Failed: The visualization is unreadable or irrelevant.
\end{itemize}

\textbf{Output Requirements:}  
\begin{itemize}
    \item Ensure the final output is in a  valid JSON format without additional text.
\end{itemize}

\textbf{Expected JSON Output Format:}
\begin{center}
\{
    "Answer Match": "...",
    "Readability and Quality Score": "...",
    "Chart Correctness Score": "..."
\}
\end{center}
    \end{minipage} \\

    \bottomrule
  \end{tabularx}
\end{table*}